\documentclass[10pt,twocolumn,letterpaper]{article}

\usepackage{cvpr}
\usepackage{times}
\usepackage{epsfig}
\usepackage{graphicx}
\usepackage{amsmath}
\usepackage{amssymb}
\usepackage{caption}
\usepackage{bm}
\usepackage{dsfont}
\usepackage{xcolor}
\usepackage{mathtools}
\usepackage{multirow}
\usepackage{cite}
\usepackage{pdfpages}

\newcommand\norm[1]{\left\lVert#1\right\rVert}

\definecolor{citecolor}{HTML}{0071bc}
\usepackage[pagebackref=false,breaklinks=true,letterpaper=true,colorlinks,citecolor=citecolor,bookmarks=false]{hyperref}
\cvprfinalcopy %

\def\cvprPaperID{890} %

\title{Learning Generative Models of Shape Handles}

\author{Matheus Gadelha \textsuperscript{1} \quad Giorgio Gori\textsuperscript{2} 
\quad Duygu Ceylan\textsuperscript{2} \quad Radom\'{\i}r M\v{e}ch\textsuperscript{2} 
\quad Nathan Carr\textsuperscript{2} \\
Tamy Boubekeur\textsuperscript{2} 
\quad Rui Wang\textsuperscript{1}
\quad Subhransu Maji\textsuperscript{1}\\
\\
\textsuperscript{1}UMass Amherst \quad \textsuperscript{2}Adobe Research
}

\DeclareMathOperator*{\argmin}{argmin}

\begin{document}

\twocolumn[{%
\renewcommand\twocolumn[1][]{#1}%
\maketitle 
  \centering
  \includegraphics[width=0.33\linewidth]{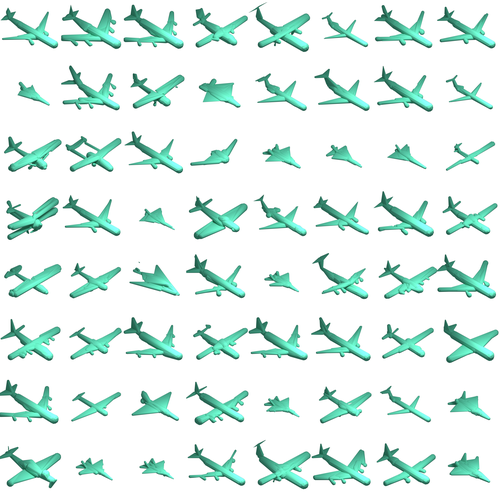}
  \includegraphics[width=0.33\linewidth]{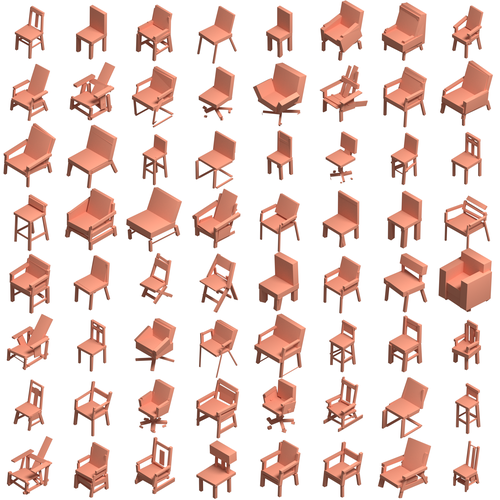}
  \includegraphics[width=0.33\linewidth]{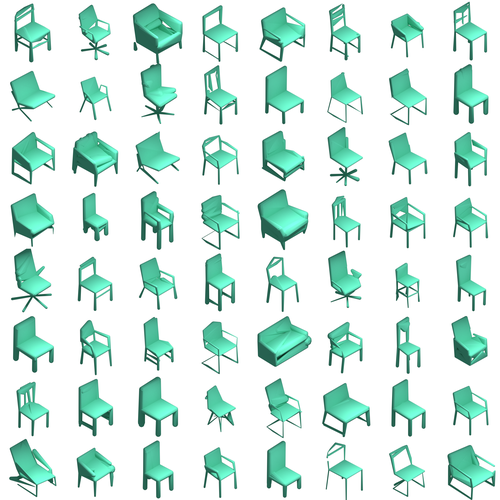}
    \vspace{-20pt}
  \captionof{figure}{\label{fig:bunny} 
  \textbf{Gallery of 3D shapes generated as sets of handles \emph{(zoom for details)}.}
  We propose a class of generative models for synthesizing sets of handles --
  lightweight proxies that can be easily utilized for high-level tasks such as shape editing, parsing, animation etc.
  Our model can generate sets with different cardinality and is flexible to work with various types of handles, such as sphere-mesh handles~\cite{spheremesh} (first and third figures) and 
  cuboids (middle figure).
}
\vspace{0.10in}
}]
\begin{abstract}
\vspace{-12pt}
We present a generative model to synthesize 3D shapes as sets of handles -- lightweight proxies that approximate the original 3D shape -- for applications in interactive editing, shape parsing, and building compact 3D representations. 
Our model can generate handle sets with varying cardinality and different types of handles (Figure~\ref{fig:bunny}).
Key to our approach is a deep architecture that predicts both the parameters and existence of shape handles, and a novel similarity measure that can easily accommodate different types of handles, such as cuboids or sphere-meshes. 
We leverage the recent advances in semantic 3D annotation as well as automatic shape summarizing techniques to supervise our approach. 
We show that the resulting shape representations are intuitive and achieve superior quality than previous state-of-the-art.
Finally, we demonstrate how our method can be used in applications such as interactive shape editing, completion, and interpolation, leveraging the latent space learned by our model to guide these tasks.
Project page: {\footnotesize \url{http://mgadelha.me/shapehandles}}.

\end{abstract}

\def\remark#1{\textcolor{red}{{C:}{#1}}}

\section{Introduction}
\label{sec:introduction}
Dramatic improvements in quality of image generation have become a key driving force behind many novel image editing applications.
Yet, similar approaches are lacking for editing and generating 3D shapes.
There are two related challenges.
First, learning generative models for 3D data is challenging, as unlike images, high-quality 3D data is hard to obtain and the data is high dimensional and often unstructured.
Second, regardless of whether good generative models are available,
manipulating and editing 3D shapes in interactive applications is harder to users than editing images.
For this reason, the geometry processing community has developed techniques for representing
3D data as a small collection of simpler \emph{proxy shapes}~\cite{ocd, acd, meshsegprim, abstractionshapes, variationalshape, obbcage, bmesh}.
In this paper, we refer to these light-weight proxies as \emph{shape handles} due to their ability to be easily manipulated by users.
These representations have been widely used in tasks that require interaction and high-level
reasoning in 3D environments, such as shape editing~\cite{gsmc_iwires_sig_09, spheremesh}, 
animation~\cite{animatedsm}, grasping~\cite{graspplaning}, and tracking~\cite{smhandtrack}.

We propose a generative models of shape handles. %
Our method adopts a two-branch network architecture to generate shapes with varying number of handles, where one branch focuses on generating handles while the other predicts the existence of each handle
(Section~\ref{sec:cardinality}). 
Furthermore, we propose a novel similarity measure based on distance fields to compare shape handle pairs. 
This measure can be easily adapted to accommodate various type of handles, such as cuboids and sphere-meshes~\cite{spheremesh}
(Section~\ref{sec:similarity}). 
Finally, in contrast to previous work~\cite{Tulsiani2017, Paschalidou2019} which focuses on unsupervised methods, we leverage recent works in collecting 3D annotations~\cite{partnet} as well as shape summarization techniques~\cite{spheremesh} to provide supervision to our approach.
Experiments show that our method significantly outperforms previous methods
on shape parsing and generation tasks.
Using self-supervised training data generated by~\cite{spheremesh}, our approach produces shapes that are twice as accurate as competing approaches in terms of intersection-over-union (IoU) metric. 
By employing human annotated data, our model can be further improved, achieving even higher accuracy than using self-supervised training data.
Moreover, as shape handles provide a compact representation, our generative networks are compact (less than 10MB).
Despite the small memory footprint,
our method generates a diverse set of
high quality 3D shapes, as seen in Figure~\ref{fig:bunny}.

\begin{figure}
\centering
\includegraphics[width=1.0\linewidth]{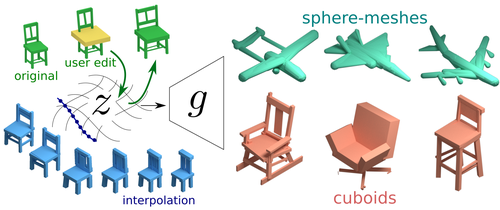}
\caption{\label{fig:handles}
\textbf{Overview}.
We propose a method to train a generative model $g$ for sets of shape handles.
Once trained, the latent representation $z$ can also be used in applications like
shape editing and interpolation.}
\vspace{-14pt}
\end{figure}

Finally, our method is built towards generating shapes using
representations that are amenable to manipulation by users.
In contrast to point clouds and other 3D representations such as occupancy grids,
handles are intuitive to modify and naturally suitable for editing and animation tasks.
The latent space of shape handles induced by the learned generative model can be leveraged to support shape editing, completion, and interpolation tasks, as depicted in Figure~\ref{fig:handles}.

\section{Related work}
\label{sec:related}

\paragraph{Deep generative models of 3D shapes.}
Multiple 3D shape representations have been used in the context of
deep generative models.
3D voxel grids~\cite{choy20163d, prgan} are a natural extension to image-based architectures,
but suffer from high memory footprint requirements.
Sparse occupancy grids~\cite{Wang-2017-ocnn, tatarchenko2017octree, hie3dcnn, matryoshka} alleviate this issue using a hierarchical grid,
but they are still not able to generate detailed shapes and they require additional bookkeeping.
Multi-view representations ~\cite{Soltani17, LunGKMW17}, point clouds~\cite{fan2016point, mrt18, pcagan, latentpc}, mesh deformations~\cite{pixel2mesh, cmrKanazawa18}
and implicit functions~\cite{park2019deepsdf, mescheder2019occupancy, chen2019learning, genova2019learning} provide alternatives that are compact and capable
of generating detailed shapes.
However, these approaches are focused on reconstructing accurate 3D shapes and
are not amenable to tasks like editing.
Our goal is different: we explore generative models to produce sets of handles --
summarized shape representations that can be easily manipulated by users.

Two closely related methods to ours are Tulsiani et al.~\cite{Tulsiani2017} and Paschalidou et al.~\cite{Paschalidou2019} where they propose models
to generate shapes as a collection of primitives \emph{without supervision}.
In contrast, we are focused
on creating models capable of utilizing shape decompositions provided by external
agents; either a human annotator or a shape summarization technique.
We demonstrate that, by using the extra information provided by annotations or well known
geometric processing techniques, our method is capable of generating
more accurate shapes while keeping the representation interpretable and intuitive for easy
editing.
Other approaches focused on learning shape structures through stronger 
supervision~\cite{li_sig17, mo2019structurenet, im2struct}, requiring not only handle description,
but also relationships between them, e.g. support, symmetry, adjacency, hierarchy, etc.
In contrast, our method models shapes as sets and we show that inter-handle relationships
can be learned directly from data, so that the latent space induced by our model can be used to guide shape editing, completion, and interpolation tasks.
Furthermore, we present a general framework that can be easily adapted to different types of handles,
not only a single parametric family, like cuboids~\cite{Tulsiani2017, mo2019structurenet, li_sig17} or superquadrics~\cite{Paschalidou2019}.

\paragraph{Methods for shape decomposition.}
Shape decomposition has been extensively studied by the geometry processing
literature. The task is to approximate a complex shape as a set of simpler, lower-dimensional
parts that are amenable for editing. We refer to these parts as \emph{shape handles}. Early cognitive studies have shown that humans tend to reason about 3D shapes as a union
of convex components~\cite{partsrecognition}. Multiple approaches have explored decomposing shapes in this manner~\cite{acd, minimumncd, acanalysis}.
However, those approaches are likely to generate too many parts, making them difficult to manipulate.
This problem was addressed by later shape approximation methods such as cages~\cite{obbcage}, 3D curves~\cite{gsmc_iwires_sig_09,mzlsgm_abstraction_siga_09,Gori2017}
and sphere-meshes~\cite{spheremesh}, which are shown very useful in shape editing and other high-level tasks. Our method is flexible to work with various types of shape handles, and in particular we show experiments using cuboids as well as sphere-meshes.

Several closely related methods to ours approximate complex shapes using
primitives such as cylinders~\cite{gdc} or cuboids~\cite{obbcage}.
These approximations are easy to interpret and manipulate
by humans. However, most existing methods rely solely on geometric cues for computing primitives, which
can lead to counter-intuitive decompositions.
In contrast, our method takes supervision from semantic information provided
by human annotators or shape summarization techniques, 
and therefore our results more accurately match human intuition.

\section{Method}
\label{sec:method}
Consider a dataset $\mathcal{D} = \{S_i\}_{i=1}^{n}$ containing $n$ sets of
shape handles.
Each set of handles $S_i$ represents a 3D shape and consists of multiple handle descriptors.
Our goal is to train a model $f_\theta$ capable of generating sets similar to the ones in
$\mathcal{D}$, i.e., using them as supervision.
More precisely, given an input $x_i$ associated with a set of handles $S_i$, our goal is to
estimate the parameters $\theta$ such that $f_\theta(x_i) \approx S_i$.
The input $x_i$ can be an image, a point cloud, an occupancy grid, or even the set of
handles itself.
When $x_i=S_i$, $f_\theta$ corresponds to an autoencoder.
If we add a regularization term to the bottleneck of $f_\theta$,
we have a Variational Auto-Encoder (VAE), which we use for
applications like shape editing, completion and interpolation (Section~\ref{sec:applications}).
However, we need to use a loss function capable of measuring the
similarity between two sets of handles, \ie the reconstruction component of a VAE.
Ideally, this loss function would be versatile -- we should be able to use it
to generate different types of handles with minimal modifications.
Moreover, our model needs to be capable of generating sets with different cardinalities,
since the sets $S_i$ do not always have the same size -- in practice,
the size of the sets used as supervision can vary a lot and our network must accommodate this need.

In this section, we describe how to create a model satisfying these constraints.
First, we describe how to compute similarities between handles.
Our method is flexible and only relies on the ability to efficiently compute the
distance from an arbitrary point in space to the handle's surface.
We then demonstrate how to use this framework with two types of handles:
cuboids and sphere-meshes.
Finally, we describe how to build models capable of generating sets with varying sizes, 
by employing a separate network branch to predict the existence of shape handles.

\subsection{Similarity between shape handles}
\label{sec:similarity}

Consider two sets of shape handles of the same type: $A=\{a_j\}_{j=1}^{|A|}$ and $B=\{b_k\}_{k=1}^{|B|}$, where
$a_j$ and $b_k$ are parameters that describe each handle.
For example, if the handle type is cuboid, $a_j$ (or $b_k$) would include the cuboid dimensions,
rotation and translation in space.  %
One way to compute similarity between sets is through Chamfer distance.
Let the asymmetric Chamfer distance between the two sets of
handles $A$ and $B$ be defined as:
\begin{equation}
    \label{eq:ch}
    Ch(A, B) = \frac{1}{|A|}\sum_{a \in A} \min_{b \in B} D(a, b)
\end{equation}
where $D(a,b)$ is a function that computes the similarity between two handles with
parameters $a$ and $b$.
There are multiple choices for $D(a,b)$.
One straightforward choice is to define $D$ as the $\ell_p$-norm of the vector $a-b$.
However, this is a poor choice as the parameters are not homogeneous. For example, parameters that describe rotations should
not contribute to the similarity metric in the same way as those describing translations.
Furthermore, there may be multiple configurations that describe the same shape -- 
e.g., vertices that are in different orders may describe the same triangle; a cube can be rotated and translated
in multiple ways and end up occupying the same region in space.

We address these problems by proposing a novel distance metric $D(a,b)$ which measures the similarity of the distance field functions of the two handles.
Specifically, let $\mathcal{P}$ be a set of points in the 3D space and let $\mu({a})$ represent the surface of the handle described by $a$.
Now, we define $D$ as follows:
\begin{equation}
    \label{eq:D}
    D(a,b) = \sum_{p \in \mathcal{P}} \Big(\min_{p_a \in \mu(a)}\norm{p-p_a}_2 - \min_{p_b \in \mu(b)}\norm{p-p_b}_2\Big)^2
\end{equation}
Intuitively, %
$D$ calculates the sum of squared differences between two
feature vectors representing the distance fields with respect to each of the handles.
Each dimension of these feature vectors contains the distance between a point in
a set of \emph{probe points} $\mathcal{P}$ and the surface of the handle 
defined by its parameters ($a$ and $b$ in Equation~\ref{eq:D}).
The main advantage of this similarity computation is its versatility:
it allows us to compare any types of shape handles; the only requirement
is the ability to efficiently compute $\min_{p_h \in \mu(h)}\norm{p-p_h}_2$ given
handle parameters $h$ and a point $p$.
In the following subsections, we show how to efficiently perform this computation
for two types of shape handles: cuboids and sphere-meshes.

\paragraph{Cuboids.}
We choose to represent a cuboid by parameters
$h = \langle\bf c, \bf l , \bf r_1, \bf r_2\rangle$, where $\mathbf{c} \in \mathbb{R}^3$ is the cuboid center, $\mathbf{l} \in \mathbb{R}^3$ is the cuboid scale factor (i.e. dimensions), $\mathbf{r_1}, \mathbf{r_2} \in \mathbb{R}^3$ are vectors describing the
rotation of the cuboid.
This rotation representation has continuity properties that benefit
its estimation through neural networks~\cite{nnrotation}.
Notice that we can build a rotation matrix $\bf R$ from $\bf r_1$ and $\bf r_2$
by following the procedure described in~\cite{nnrotation}.
Now, consider the transformation $\tau_h(p) = \mathbf{R}^Tp - \mathbf{c}$.
Let $\mu_C(h) \in \mathbb{R}^3$ represent the surface of the cuboid parametrized by $h$. We can compute $\min_{p_h \in \mu_C(h)}\norm{p-p_h}_2$
(i.e. distance from $p$ to the cuboid) as follows: %
$$
\min_{p_h \in \mu_C(h)}\norm{p-p_h}_2 = 
\norm{(|\tau_h(p)| - \bf l)^+}_2 + \big(\max(|\tau_h(p)| - \mathbf{l})\big)^-
$$
where $(\cdot)^+$, $(\cdot)^-$ and $|\cdot|$ represent element-wise $\max(\cdot, 0)$, $\min(\cdot, 0)$ and absolute value, respectively.
Since this expression can be computed in $O(1)$, 
we are able to compute Equation~\ref{eq:D} in $O(|\mathcal{P}|)$, where the number of probing points $|\mathcal{P}|$ is relatively small.
In practice, we sample 64 points in a regular grid
in the unit cube.

\paragraph{Sphere-meshes.}

\begin{figure}
\centering
\includegraphics[width=1.0\linewidth]{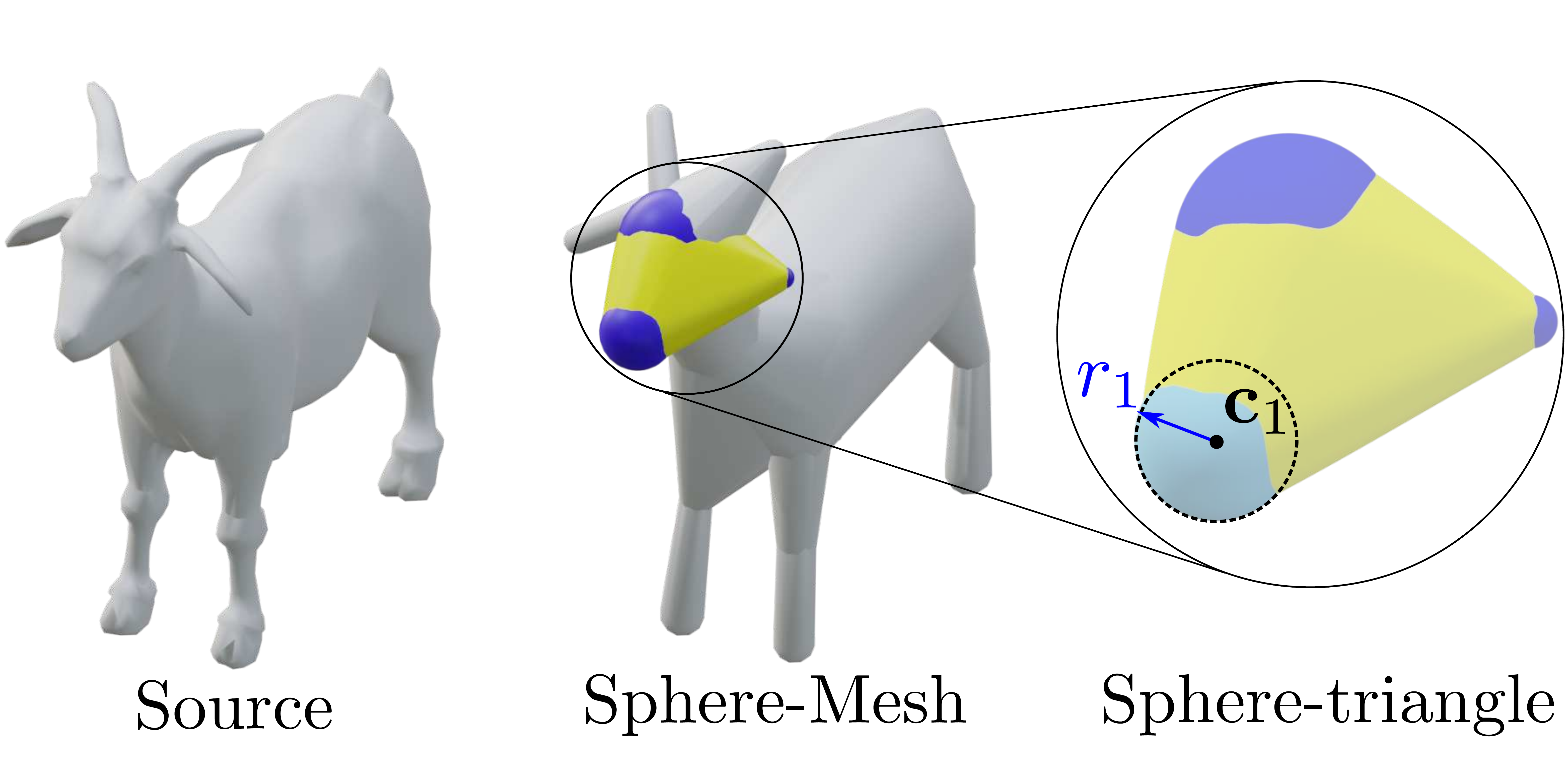}
\caption{\label{fig:spheremesh} \small
Schematic representation of sphere-meshes.
A sphere-mesh (middle) is computed from a regular triangle mesh (left) as input, and it consists of multiple sphere-triangles (right), each of which is a volumetric representation}
\end{figure}

A triangle mesh consists of a set of vertices and triangular faces representing
the vertex connectivity.
Every vertex is a point in space and the surface of a triangle contains all the points
that can be generated by interpolating the triangle's vertices using
barycentric coordinates.
A sphere-mesh is a generalization of a triangle mesh -- every vertex is
a sphere instead of a point in space.
Thus, every sphere-mesh ``triangle" is actually a volume delimited by the convex-hull
of the spheres centered at the triangle vertices.
Figure~\ref{fig:spheremesh} presents a visual description of sphere-mesh components.
Thiery et al.~\cite{spheremesh} introduced an algorithm to compute sphere-meshes from
regular triangle meshes.
They show that complex meshes can be closely approximated with a sphere-mesh containing
a fraction of the original components.

We model sphere-meshes as a set of sphere-mesh triangles, called \emph{sphere-triangles}.
Similarly to a regular triangle, a sphere-triangle is fully defined by 
its vertices, the difference being that its vertices are now spheres instead of
points.
Thus, we choose to represent a sphere-triangle using parameters
$h = \langle r_1, r_2, r_3, \bf c_1, c_2, c_3 \rangle$; where 
$\mathbf{c_1}, \mathbf{c_2}, \mathbf{c_3} \in \mathbb{R}^3$
are the centers of the three spheres, and $r_1, r_2, r_3 \in \mathbb{R}^+$ are
their radii.
Let $\mu_T(h)$ represent the the surface of the sphere-triangle parametrized by $h$.
For calculating the similarity between two sphere-triangles: as each sphere-triangle is uniquely defined by its three spheres, it suffices to have $\mu_T$ contain only the surfaces of these three spheres, and hence it does not need to contain the entire sphere triangle.
Thus, the distance of a probing point $p$ to the handle surface is computed as follows:
$$
\min_{p_h \in \mu_T(h)}\norm{p-p_h}_2 = 
\min_{i\in\{1, 2, 3\}} (\norm{p - \mathbf{c}_i}_2 - r_i).
$$

\subsection{Generating sets with varying cardinality}
\label{sec:cardinality}
The neural network $f$ generates shapes represented by sets of handles given an input $x$.
Our design of $f$ includes two main components: an encoder $q$ that, given an input $x$, outputs a latent set representation
$z$; and a decoder $g$ that, given the latent set representation $z$, generates
a set of handles.
Even though we can use a symmetric version of Equation~\ref{eq:ch} 
to compute the similarity between the generated set $g(q(x_i))$ 
and the ground-truth set of handles $S_i$, 
so far our model has not taken into account the varying size (i.e. number of elements) of the generated sets.
We address this issue by separating the generator into two parts: a parameter prediction
branch $g_p$ and an existence prediction branch $g_e$. The parameter prediction branch is trained to always output a fixed number of handle parameters where
$[g_p(z)]_i$ represents the parameters of the $i^{th}$ handle. On the other hand, the existence prediction branch $[g_e(z)]_i \in [0,1]$ represents the probability of existence of the $i^{th}$
generated handle.
Now, we need to adapt our loss function to consider the probability of
existence of a handle.

If we assume that all handles exist, our model can be trained using the following
loss: %
$$
\mathcal{L} = Ch(g_p(z_i), S_i) + Ch(S_i, g_p(z_i)),
$$
where $S_i$ is a set of shape handles drawn from the training data and
$z_i$ is a latent representation computed from the associated input $x_i$.
However, we want to modify this loss to take into account the probability of a handle existing
or not. To do so, note that $\mathcal{L}$ has two terms.
The first term measures accuracy: i.e. how close each of the handles in $g_p(z_i)$ is from the handles
in $S_i$.
For this term, we can use $g_e$ as weights for the summation in Equation~\ref{eq:ch},
which leads to the following definition:
\begin{equation}
\label{eq:prec}
P(z, S) = \sum_{i=1}^K \min_{s \in S} D([g_p(z)]_i, s)[g_e(z)]_i,
\end{equation}
where $z$ is a latent space representation,  $S$ is a set of handles and
$K = |g_p(z)| = |g_e(z)|$.
The intuition is quite simple:
if the $i^{th}$ handle is likely to exist, its distance to the closest handle should be taken into consideration;
on the other hand, if the $i^{th}$ handle is unlikely to exist, it does not matter if
it is approximating a handle in $S$ or not.

The second term in $\mathcal{L}$ measures coverage:
every handle in $S_i$ must have (at least) one handle in the generated set that is very
similar to it.
Here, we use an insight presented in~\cite{Paschalidou2019} to efficiently compute the coverage of
$S_i$ while considering the probability of elements in a set existing or not.
Let $g_p^s(z)$ be the list of generated handles $g_p(z)$ ordered in non-decreasing
order according to $D([g_p^s]_i, s)$ for $i=1,...,|g_p(z)|$.
We compute the coverage of a set $S$ from a set generated from $z$ as follows:
\begin{equation}
    \label{eq:cov}
C(z, S) = \sum_{s \in |S|} \sum_{i=1}^{K}D([g_p^s(z)]_i, s)[g_e^s(z)]_i \prod_{j=1}^{i-1}(1 - [g_e^s(z)]_j).
\end{equation}
The idea behind this computation is the following:
for every handle $s \in S$, we compute its distance to every handle in $g_p(z)$, weighted
by the probability of that handle existing or not.
However, the distance to a specific handle is important only if no other handle closer to $s$ exists. Thus, the whole term needs to be weighted by $\prod_{j=1}^{i}(1 - [g_e^s(z)]_j)$.
Finally, we can combine Equations~\ref{eq:prec} and \ref{eq:cov} to define the
reconstruction loss $\mathcal{L}_{rec}$ used to train our model:
\begin{equation}
    \label{eq:rec}
    \mathcal{L}_{rec} = P(z, S) + C(z, S).
\end{equation}

\paragraph{Alternate training procedure.}
\begin{figure}
\centering
\includegraphics[width=1.0\linewidth]{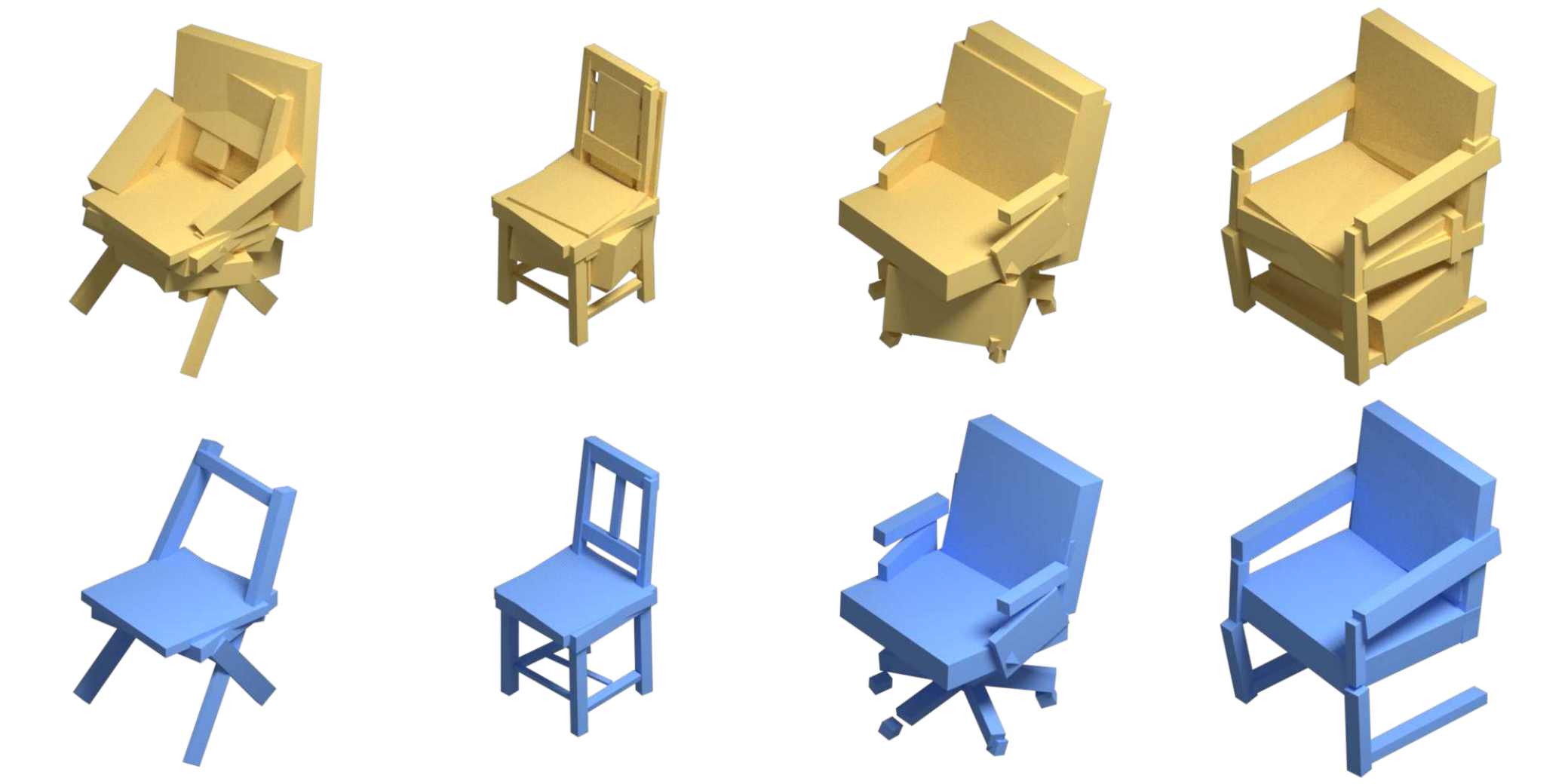}
\caption{\label{fig:alttrain} \small
Comparison of results after the first stage (top row) and second
stage (bottom row) of alternate training.
While the first stage ensures coverage, some extra, unnecessary handles are also generated.
The second stage trains the existence branch, which assigns a low probability
of existence to the inaccurate handles.}
\end{figure}

Although minimizing the loss in Equation~$\ref{eq:rec}$ at once enables
generating sets of different sizes, our experiments show that the
results can be further improved if we train $g_p$ and $g_e$ in an alternating fashion.
Specifically, we first initialize the biases and weights of the last layer of $g_e$ to ensure
that all of its outputs are $1$, i.e., the model is initialized to predict that every primitive exists.
Then, in the first stage of the training, we fix the parameters of $g_e$ and train $g_p$ minimizing only the coverage
$C(z, S)$.
During the second stage of the training, we fix the parameters of $g_p$ and
update the parameters of $g_e$, but this time minimizing the full reconstruction loss
$\mathcal{L}_{rec}$.
As we show in Section 4, this alternating procedure improves the training leading to the generation of more
accurate shape handles.  
The intuition is that while training the model to predict the handle parameters ($g_p$), the
network should be only concerned about coverage, i.e., generating at least one similar handle for each 
ground-truth handle.
On the other hand, while training the existence prediction branch ($g_e$), we want
to remove the handles that are in incorrect positions while keeping the coverage of the ground-truth set. %

\section{Experiments}
\label{sec:experiments}

This section describes our experiments and validates results.
We experimented with two different types of handles:
cuboids computed from PartNet~\cite{partnet} segmentations and
sphere-meshes computed from ShapeNet~\cite{shapenet} shapes using~\cite{spheremesh}.
We compare our results to two other approaches focused on generating shapes
as a set of simple primitives, namely cuboids~\cite{Tulsiani2017} 
and superquadrics~\cite{Paschalidou2019}.
All the experiments in the paper were implemented using Python 3.6
and PyTorch.
Computation was performed on TitanX GPUs.

\subsection{Datasets}
\paragraph{Cuboids from PartNet~\cite{partnet}.}
We experiment with human annotated handles by fitting cuboids to the parts
segmented in PartNet~\cite{partnet}.
The dataset contains 26,671 shapes from 24 categories and 573,585 part instances.
In order to compare our model with other approaches trained
on the ShapeNet~\cite{shapenet} chairs dataset, we select the subset of PartNet chairs
that is also present in ShapeNet.
This results in 6773 chair models segmented in multiple parts.
Every model has on average 18 parts, but there are also examples with as many as 137 parts.
For every part we fit a corresponding cuboid using PCA. 
Then, we compute the volume of every cuboid and keep at topmost 30 cuboids in terms of volume.
Notice that 92\% of the shapes have less than 30 cuboids, so those remain unchanged.
The others will have missing components, but those usually correspond to very small details
and can be ignored without degrading the overall structure.

\paragraph{Sphere-meshes from ShapeNet~\cite{shapenet}.}
In contrast to cuboids (which are computed from human annotated parts), we compute sphere-meshes fully automatically using the procedure
described in~\cite{spheremesh}.
We use ShapeNet categories that are also analyzed in~\cite{Paschalidou2019, Tulsiani2017}:
chairs, airplanes and animals.
The sphere-mesh computation procedure requires pre-selecting how many sphere-vertices to use.
The algorithm starts by considering the regular triangle mesh as a trivial sphere-mesh (vertices with null radius) and then decimates the
original mesh progressively through edge collapsing, optimizing for new sphere-vertex each time an edge is removed.
This procedure is iterated until the required number of vertices is achieved.

In our case, since our model is capable of generating sets with different cardinalities,
we are not required to set a fixed number of primitives for every shape. Therefore we use the following method to compute a sphere-mesh with adaptive number of vertices.
Specifically, for every shape in the dataset, we start by computing a sphere-mesh with
10 vertices. 
Then, we sample 10K points both on the sphere-mesh surface and the original mesh.
If the Hausdorff distance between the point clouds is smaller than $\epsilon=0.2$
(point clouds are normalized to fit the unit sphere), we keep the current computed sphere-mesh.
Otherwise, we compute a new sphere-mesh by incrementing the number of vertices.
This procedure continues until we reach a maximum of 40 vertices.
This adaptive sphere-mesh computation allows our model to achieve
a good balance between shape complexity and summarization --
simpler shapes will be naturally represented with a smaller number of primitives.
We note that the sphere-mesh computation allows the resulting mesh to contain not only triangles, but also edges (i.e. degenerate triangles).
For simplicity, we make no distinction between sphere-triangles or edges:
edges are simply triangles that have two identical vertices.

\begin{figure}[t]
\centering
\includegraphics[width=1.0\linewidth]{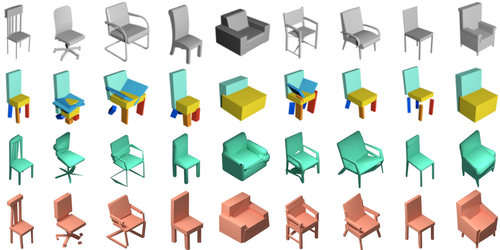}
\caption{\label{fig:chairs} \small
\textbf{Shape parsing on the chairs dataset}. From top to bottom, we show
ground-truth shapes,
results by Tulsiani et al.~\cite{Tulsiani2017}, results by our method using sphere-mesh handles, and our method using cuboids handles. 
Note how our results (last two rows) are able to generate handles with much better details such as the stripes on the back of the chair (first column),
legs on wheel chairs (second column) and armrests in several other columns.
}
\vspace{-10pt}
\end{figure}

\subsection{Shape Parsing}

The shape parsing task is to compute a small set of primitives from non-parsimonious,
raw, 3D representations, like occupancy grids, meshes or point clouds.
We analyze the ability of our model in performing shape parsing using a similar setup
to~\cite{Paschalidou2019, Tulsiani2017}.
Specifically, following the notation defined in Section~\ref{sec:method},
we train a model $f_\theta$ using input-output pairs $\langle x_i, S_i \rangle$, where
$x_i$ corresponds to a point cloud with 1024 points and $S_i$ is a set of handles
summarizing the shape represented by $x_i$.
We use a PointNet~\cite{pointnet} encoder to process
a point cloud with 1024 points and generate a 1024 dimensional encoding.
This encoding is then used as an input for our two-branched set decoder.
Both branches follow the same architecture: 3 fully connected layers with 256 hidden neurons
followed by batch normalization and ReLU activations.
The only difference between the two branches is in the last layer.
Assume $N$ is the maximum set cardinality generated by our model and $D$ is the handle
dimensionality (i.e. number of parameters of each handle descriptor, which happens to be $D=12$ for both sphere-mesh and cuboid). Then
$g_p$ outputs $N\times D$ values followed by a $\tanh$ activation, while $g_e$
outputs $N$ values followed by a sigmoid activation.
We set $N=30$ for cuboid handles and $N=50$ for sphere-meshes. %
The model is trained end-to-end by using the alternating training described in Section~\ref{sec:method}.
Training is performed using the Adam optimizer with a learning rate of $10^{-3}$ for 5K iterations
in each stage.

\begin{figure}
\centering
\includegraphics[width=1.0\linewidth]{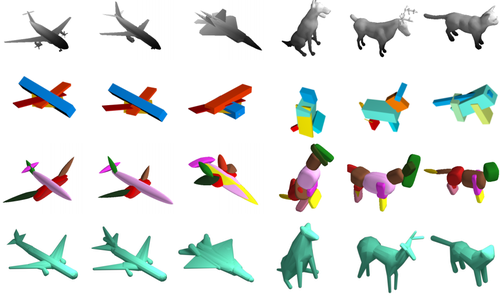}
\vspace{-20pt}
\caption{\label{fig:all} \small
\textbf{Shape parsing on the airplanes and animals datasets}. From top to bottom, we show ground-truth shapes,
results by Tulsiani et al.~\cite{Tulsiani2017},
results by Paschalidou et al.~\cite{Paschalidou2019}, and results by our model trained using sphere-mesh handles.
Our results contain accurate geometric details, such as
the engines on the airplanes and animal legs that are clearly separated.
}
\end{figure}

Figures~\ref{fig:chairs} and \ref{fig:all} show visual comparisons of our method with previous work. Qualitatively, our method generates shape handles with accurate geometric details, including many thin structures that previous methods struggle with. 

\begin{table}[t]
\centering
\small
\begin{tabular}{l|l|ccc}
                   & \multirow{2}{*}{\textbf{Handle type}}  & \multicolumn{3}{c}{\textbf{Category}}        \\
                   &              & Chairs   & Airplanes & Animals \\
                   \hline
\cite{Tulsiani2017}    & Cuboid       & 0.129   & 0.065    & 0.334  \\
\cite{Paschalidou2019} & Superquadric & 0.141   & 0.181    & 0.751  \\
\hline
\multirow{ 2}{*}{Ours}               & Cuboid       & 0.311    & -         & -       \\
               & Sphere-mesh  & \textbf{0.298}    & \textbf{0.307}     & \textbf{0.761}      \\
\hline
\end{tabular}
\caption{
\label{tab:parse}
\textbf{Quantitative results for shape parsing.} Intersection over union computed
on the reconstructed shapes. The best self-supervised results are shown in bold font.
}
\end{table}

\paragraph{Quantitative comparisons}

We compare our method against~\cite{Tulsiani2017, Paschalidou2019} using intersection
over union (IoU) metric and results are shown in Table~\ref{tab:parse}.
As expected, when using cuboids as handles, our method leverages the annotated data from the PartNet~\cite{partnet} to achieve significantly more accurate shape
approximations (more than twice the IoU in comparison).
On the other hand, as~\cite{Tulsiani2017, Paschalidou2019} are trained without leveraging annotated data, a more fair comparison is between theirs and our method using sphere-mesh handles, which are computed automatically.
Our method still clearly outperforms theirs in all categories --
chairs, airplanes and animals.
This shows that even though a neural network in theory should be able to learn the best parsimonious
shape representations, using self-supervision generated by shape summarization techniques (e.g. sphere-meshes) can still help it achieve more accurate approximations. 

\begin{figure}[t]
\centering
\includegraphics[width=1.0\linewidth]{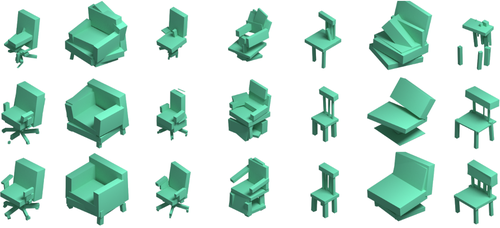}
\vspace{-15pt}
\caption{\label{fig:abl} \small
\textbf{Ablation studies.}
Shapes generated from a model trained without our proposed handle similarity metric (first row),
model trained without the two-stage training procedure (second row), and our full model (last row).
Note that comparing handles using just $\ell_2$-norm (first row) yields poor results.
Training $g_p$ and $g_e$ at the same time (instead of alternating) yields reasonable results, but some parts
are missing and/or poorly oriented.
}
\end{figure}
\begin{table}[]
\centering
\begin{tabular}{ccc}
\hline
w/o similarity & w/o alternate & full model \\
0.192 & 0.320 & \textbf{0.352}\\
\hline
\end{tabular}
\caption{ \small
\label{tab:abl}
Quantitative results of ablation studies comparing our full model with two variations that lack our handle similarity metric and alternate training procedure respectively.
}
\vspace{-0.1in}
\end{table}

\subsection{Ablation studies}
\label{sec:ablation}
We investigated the influence of the two main contributions of this work:
the similarity metric for handles and the alternating training procedure for $g_p$ and $g_e$.
To do so, we adopt a shape-handle auto-encoder and compare different variations by computing
the IoU of reconstructed shapes in a held-out test set.
The auto-encoder architecture is very similar to the one used in shape parsing, except for
the encoder -- it still follows a PointNet architecture, but every ``point'' is actually
a handle treated as a point in a $D$-dimensional space.
We analyzed three different variations.
In the first one, we simply used the $\ell_2$-norm between the handle parameters (cuboids, in this case).
As shown in Figure~\ref{fig:abl} and Table~\ref{tab:abl},
the proposed handle similarity metric has a significant impact on the quality of the
generated shapes.
The second variation consists of training the same model, but without using the alternating procedure
described in Section~\ref{sec:method}. %
Figure~\ref{fig:abl} shows that the alternating training procedure generates
more accurate shapes, with fewer missing parts and better cuboid orientation.

\subsection{Applications}
\label{sec:applications}
In this section, we demonstrate the use of our generative model in several applications.
We employed a Variational Auto-Encoder (VAE)~\cite{vae} for this purpose.
It follows the same architecture as the auto-encoder described in Section~\ref{sec:ablation} with the 
only difference being that the output of the encoder (latent representation $z$) has
dimensionality 256 instead of 512.
Additionally, following~\cite{mrt18}, we added an additional regularization term to the training objective:
\begin{equation}
    \mathcal{L}_{reg} = \norm{cov(Q(x) + \delta)}_2 + \mathbb{E}_{x\sim\mathcal{D}}[Q(x)]
\end{equation}
where $Q$ is the encoder, $cov(\cdot)$ is the covariance matrix, $\norm{\cdot}_2$ is the Frobenius norm, $x$ is input handle set and $\delta$ is random noise sampled from $\mathcal{N}(0,cI)$.
Thus, the network is trained minimizing the following function:
\begin{equation}
    \mathcal{L} = \mathcal{L}_{rec} + \lambda\mathcal{L}_{reg}.
\end{equation}
In all our experiments, we used $\lambda=0.1$ and $c=0.01$.
The model is trained using the alternate procedure described before, \ie $\mathcal{L}_{rec}$ is replaced by $C(z,S)$ while training $g_p$.

\vspace{5pt}
\noindent\textbf{Interpolation.}
\begin{figure}
\centering
\includegraphics[width=1.0\linewidth]{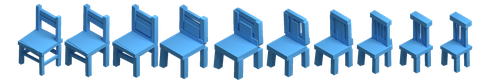}
\vspace{-20pt}
\caption{\label{fig:interp} \small
\textbf{Latent space interpolation}
Sets of handles can be interpolated by linearly interpolating
the latent representation $z$.
Transitions are smooth and generate plausible intermediate shapes.
Notice that the interpolation not only changes handle parameters, but also
adds new handles / removes existing handles as necessary.
}
\vspace{-15pt}
\end{figure}
Once the VAE model is trained, we are able to morph between two shapes by 
linearly interpolating their latent representations $z$.
In particular, we sample two values $z_1, z_2$ from $\mathcal{N}(0, I)$ and generate new shapes by passing the interpolated encodings $\alpha z_1 + (1-\alpha) z_2$ through the decoder $g$,
where $\alpha \in [0, 1]$.
Results using cuboid handles are presented in Figure~\ref{fig:interp}.
Note that the shapes are smoothly interpolated, with new handles added and old handles removed as necessary when the overall shape deforms.
Additionally, relationships between handles, like symmetries, adjacency and support, are
preserved, thanks to the latent space learned by our model, even though such characteristics are never explicitly specified as supervision.

\begin{figure}
\centering
\includegraphics[width=1.0\linewidth]{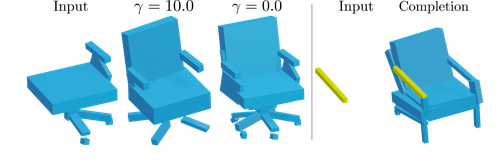}
\vspace{-18pt}
\caption{\label{fig:completion} \small
\textbf{Results of handle completion}.
Recovering full shape from incomplete set of handles.
Using $\gamma$ to control the complexity of the completed shape (left).
Predicting a complete chair from a single handle (right).
}
\vspace{-0.1in}
\end{figure}

\vspace{5pt}
\noindent\textbf{Handle completion.}
Consider an incomplete set of handles $A=\{a_i\}_{i=1}^{N}$ as input, the handle completion task is to generate a complete set of handles $A^*$,
such that $A^*$ contains not only the handles in the input $A$ but also necessary additional handles that result in a plausible shape.
For example, given a single cuboid handle as shown in Figure~\ref{fig:completion},
we want to generate a complete chair that contains that input handle.
We perform this task by finding a latent representation $z^*$ that generates a set of handles approximating the elements
in $A$. Specifically, we solve the following optimization problem:
\begin{equation}
   z^* = \argmin_{z \in \mathcal{Z}}C(z, A), \quad A^* = g(z^*),
   \label{eq:completion}
   \vspace{-5pt}
\end{equation}
where $C$ is the coverage metric defined in Equation~\ref{eq:cov} and $A^*$ is the completed shape (i.e. output of the decoder using $z^*$ as input).
We can also use the existence prediction branch ($g_e$) in this framework to reason
about how complex we want the completed shapes to be.
Specifically, we add an additional term to the optimization:
\begin{equation}
    \label{eq:completion}
   z^* = \argmin_{z \in \mathcal{Z}}C(z, A) + \gamma\sum_{i=1}^{N}[g_e(z)]_i,
   \vspace{-5pt}
\end{equation}
where $\gamma$ controls the complexity of the shape.
If $\gamma=0$, we are not penalizing a set with multiple handles -- only
coverage matters.
As $\gamma$ increases, existence of multiple handles is penalized more, leading to a solution with a lower cardinality.
As can be seen in Figure~\ref{fig:completion}, our model is capable of recovering plausible chairs even when given a single handle. In addition, we can generate multiple proposals for $A^*$ by initializing the optimization with different values of $z$. More results can be found in the supplemental material. 

\begin{figure}
\centering
\includegraphics[width=1.0\linewidth]{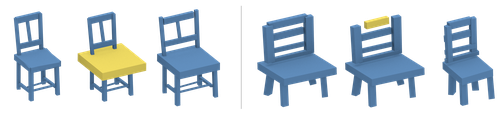}
\vspace{-20pt}
\caption{\label{fig:editing} \small
\textbf{Editing chairs}.
Given an initial set of handles, a user can modify any handle (yellow). Our model then
updates the entire set of handles, resulting in a modified shape which observes the user edits while preserving the overall structure.}
\vspace{-0.2in}
\end{figure}

\paragraph{Shape editing.}
For editing shapes, we use a similar optimization based framework.
Consider an original set of handles $A$ describing a particular shape.
Assume that the user made edits to $A$ by modifying the parameters of some handles, creating
a new set $A^\prime$.
Our goal is to generate a plausible new shape $A^*$ from $A^\prime$, while minimizing the deviation 
from the original shape.
To achieve this goal, we solve the following minimization problem via gradient descent:
\begin{equation}
   \label{eq:Editing}
   z^* = \argmin_{z \in \mathcal{Z}}C(z, A^\prime) + \gamma\norm{z - z_A}_2, \quad A^* = g(z^*)
\end{equation}
where $z_A$ is the latent representation of the original shape.
The intuition for Equation~\ref{eq:Editing} is simple: we want to generate a plausible shape that
approximates the user edits by minimizing  $C(z, A^\prime)$ but also keep the overall
characteristics of the original shape $A$ by adding a penalty for deviating
too much from $z_A$.
Results are shown in Figure~\ref{fig:editing}. 
As observed in the figure, when the user edits one of the handles, 
our model can automatically modify the shape of the entire chair while preserving its overall structure. 

\paragraph{Limitations.}
Our method has several limitations to be addressed in future work. 
First, during training we set a maximum number of handles to be generated. 
Increasing this number would allow more complex shapes but also entail a larger network with higher capacity. 
Therefore, there is a trade-off between the compactness of the generative model and the desired output complexity. 
Furthermore, our method currently does not guarantee the output handles observe certain geometric constraints, such as parts that need to be axis-aligned or orthogonal to each other. 
For man-made shapes, these are often desirable constraints and even slight deviation is immediately noticeable. 
While our model can already learn geometric relationships among handles from the data directly, generated shapes might benefit from additional supervision enforcing geometric constraints.

\section{Conclusion}
We presented a method to generate shapes represented as
sets of handles -- lightweight proxies that approximate the original
shape and are amenable to high-level tasks, like shape editing, parsing and animation.
Our approach leverages pre-defined sets of handles as supervision, either through annotated data or self-supervised methods. 
We proposed a versatile similarity metric for shape handles that can easily accommodate different types of handles, and a two-branch network architecture to generate handles with varying cardinality.
Experiments show that our model is capable of generating compact and accurate
shape approximations, outperforming previous work. We demonstrate our method in a variety of applications, including interactive shape editing, completion, and interpolation, leveraging the latent space learned by our model to guide these tasks.

\paragraph{Acknowledgements.} This work is supported in part by a gift from Adobe Research, 
NSF grants 1908669, 1749833. Our experiments were performed in the UMass GPU cluster obtained under the Collaborative Fund managed by the Massachusetts Technology Collaborative.

\clearpage
{\small
\bibliographystyle{ieee_fullname}
\bibliography{egbib}

\begin{thebibliography}{10}\itemsep=-1pt

\bibitem{latentpc}
Panos Achlioptas, Olga Diamanti, Ioannis Mitliagkas, and Leonidas~J Guibas.
\newblock {Learning Representations and Generative Models For 3D Point Clouds}.
\newblock In {\em International Conference on Machine Learning}, 2018.

\bibitem{meshsegprim}
Marco Attene, Bianca Falcidieno, and Michela Spagnuolo.
\newblock Hierarchical mesh segmentation based on fitting primitives.
\newblock {\em The Visual Computer}, 22(3):181--193, 2006.

\bibitem{shapenet}
Angel~X Chang, Thomas Funkhouser, Leonidas Guibas, Pat Hanrahan, Qixing Huang,
  Zimo Li, Silvio Savarese, Manolis Savva, Shuran Song, Hao Su, et~al.
\newblock Shapenet: An information-rich 3d model repository.
\newblock {\em arXiv preprint arXiv:1512.03012}, 2015.

\bibitem{ocd}
Bernard Chazelle and {David Paul} Dobkin.
\newblock {\em Optimal Convex Decompositions}, pages 63--133.
\newblock Number~C in Machine Intelligence and Pattern Recognition. 1985.

\bibitem{chen2019learning}
Zhiqin Chen and Hao Zhang.
\newblock Learning implicit fields for generative shape modeling.
\newblock In {\em The IEEE Conference on Computer Vision and Pattern
  Recognition}, 2019.

\bibitem{choy20163d}
Christopher~B Choy, Danfei Xu, JunYoung Gwak, Kevin Chen, and Silvio Savarese.
\newblock {3D-R2N2: A} unified approach for single and multi-view {3D} object
  reconstruction.
\newblock In {\em European Conference on Computer Vision}, 2016.

\bibitem{variationalshape}
David Cohen-Steiner, Pierre Alliez, and Mathieu Desbrun.
\newblock Variational shape approximation.
\newblock In {\em ACM SIGGRAPH 2004 Papers}, SIGGRAPH '04, New York, NY, USA,
  2004. ACM.

\bibitem{fan2016point}
Haoqiang Fan, Hao Su, and Leonidas Guibas.
\newblock {A Point Set Generation Network for 3D Object Reconstruction from a
  Single Image}.
\newblock In {\em IEEE Conference on Computer Vision and Pattern Recognition},
  2017.

\bibitem{pcagan}
Matheus Gadelha, Subhransu Maji, and Rui Wang.
\newblock 3d shape generation using spatially ordered point clouds.
\newblock In {\em British Machine Vision Conference (BMVC)}, 2017.

\bibitem{prgan}
Matheus Gadelha, Subhransu Maji, and Rui Wang.
\newblock {Unsupervised 3D Shape Induction from 2D Views of Multiple Objects}.
\newblock In {\em International Conference on 3D Vision (3DV)}, 2017.

\bibitem{mrt18}
Matheus Gadelha, Rui Wang, and Subhransu Maji.
\newblock {Multiresolution Tree Networks for 3D Point Cloud Processing}.
\newblock In {\em ECCV}, 2018.

\bibitem{gsmc_iwires_sig_09}
Ran Gal, Olga Sorkine, Niloy~J. Mitra, and Daniel Cohen-Or.
\newblock iwires: An analyze-and-edit approach to shape manipulation.
\newblock {\em {ACM} Transactions on Graphics (Siggraph)}, 28(3):\#33, 1--10,
  2009.

\bibitem{genova2019learning}
Kyle Genova, Forrester Cole, Daniel Vlasic, Aaron Sarna, William~T Freeman, and
  Thomas Funkhouser.
\newblock Learning shape templates with structured implicit functions.
\newblock In {\em International Conference on Computer Vision}, 2019.

\bibitem{Gori2017}
Giorgio Gori, Alla Sheffer, Nicholas Vining, Enrique Rosales, Nathan Carr, and
  Tao Ju.
\newblock Flowrep: Descriptive curve networks for free-form design shapes.
\newblock {\em ACM Transaction on Graphics}, 36(4), 2017.

\bibitem{hie3dcnn}
Christian H{\"{a}}ne, Shubham Tulsiani, and Jitendra Malik.
\newblock Hierarchical surface prediction for 3d object reconstruction.
\newblock In {\em International Conference on 3D Vision (3DV)}, 2017.

\bibitem{partsrecognition}
D.~D. Hoffman and W.~A. Richards.
\newblock Parts of recognition.
\newblock {\em Cognition}, 18(1-3):65--96, 1984.

\bibitem{bmesh}
Zhongping Ji, Ligang Liu, and Yigang Wang.
\newblock B-mesh: A modeling system for base meshes of 3d articulated shapes.
\newblock {\em Computer Graphics Forum}, 29(7):2169--2177, 2010.

\bibitem{acanalysis}
Oliver~Van Kaick, Noa Fish, Yanir Kleiman, Shmuel Asafi, and Daniel Cohen-OR.
\newblock Shape segmentation by approximate convexity analysis.
\newblock {\em ACM Trans. Graph.}, 34(1), 2014.

\bibitem{cmrKanazawa18}
Angjoo Kanazawa, Shubham Tulsiani, Alexei~A. Efros, and Jitendra Malik.
\newblock Learning category-specific mesh reconstruction from image
  collections.
\newblock In {\em ECCV}, 2018.

\bibitem{vae}
Diederik~P Kingma and Max Welling.
\newblock Auto-encoding variational bayes.
\newblock {\em arXiv preprint arXiv:1312.6114}, 2013.

\bibitem{li_sig17}
Jun Li, Kai Xu, Siddhartha Chaudhuri, Ersin Yumer, Hao Zhang, and Leonidas
  Guibas.
\newblock Grass: Generative recursive autoencoders for shape structures.
\newblock {\em ACM Transactions on Graphics (Proc. of SIGGRAPH 2017)}, 36(4):to
  appear, 2017.

\bibitem{acd}
Jyh-Ming Lien and Nancy~M. Amato.
\newblock Approximate convex decomposition of polyhedra.
\newblock In {\em Proceedings of the 2007 ACM Symposium on Solid and Physical
  Modeling}, SPM '07, 2007.

\bibitem{LunGKMW17}
Zhaoliang Lun, Matheus Gadelha, Evangelos Kalogerakis, Subhransu Maji, and Rui
  Wang.
\newblock 3d shape reconstruction from sketches via multi-view convolutional
  networks.
\newblock In {\em International Conference on 3D Vision (3DV)}, 2017.

\bibitem{abstractionshapes}
Ravish Mehra, Qingnan Zhou, Jeremy Long, Alla Sheffer, Amy Gooch, and Niloy~J.
  Mitra.
\newblock Abstraction of man-made shapes.
\newblock In {\em ACM SIGGRAPH Asia 2009 Papers}, SIGGRAPH Asia '09. ACM, 2009.

\bibitem{mzlsgm_abstraction_siga_09}
Ravish Mehra, Qingnan Zhou, Jeremy Long, Alla Sheffer, Amy Gooch, and Niloy~J.
  Mitra.
\newblock Abstraction of man-made shapes.
\newblock {\em {ACM} Transactions on Graphics}, 28(5):\#137, 1--10, 2009.

\bibitem{mescheder2019occupancy}
Lars Mescheder, Michael Oechsle, Michael Niemeyer, Sebastian Nowozin, and
  Andreas Geiger.
\newblock {Occupancy networks: Learning 3D reconstruction in function space}.
\newblock In {\em The IEEE Conference on Computer Vision and Pattern
  Recognition}, 2019.

\bibitem{graspplaning}
A.~T. {Miller}, S. {Knoop}, H.~I. {Christensen}, and P.~K. {Allen}.
\newblock Automatic grasp planning using shape primitives.
\newblock In {\em 2003 IEEE International Conference on Robotics and Automation
  (Cat. No.03CH37422)}, volume~2, pages 1824--1829 vol.2, 2003.

\bibitem{mo2019structurenet}
Kaichun Mo, Paul Guerrero, Li Yi, Hao Su, Peter Wonka, Niloy Mitra, and
  Leonidas Guibas.
\newblock Structurenet: Hierarchical graph networks for 3d shape generation.
\newblock {\em ACM Transactions on Graphics (TOG), Siggraph Asia 2019},
  38(6):Article 242, 2019.

\bibitem{partnet}
Kaichun Mo, Shilin Zhu, Angel~X. Chang, Li Yi, Subarna Tripathi, Leonidas~J.
  Guibas, and Hao Su.
\newblock Partnet: A large-scale benchmark for fine-grained and hierarchical
  part-level 3d object understanding.
\newblock In {\em The IEEE Conference on Computer Vision and Pattern
  Recognition (CVPR)}, June 2019.

\bibitem{im2struct}
Chengjie Niu, Jun Li, and Kai Xu.
\newblock Im2struct: Recovering 3d shape structure from a single rgb image.
\newblock In {\em Computer Vision and Pattern Regognition (CVPR)}, 2018.

\bibitem{park2019deepsdf}
Jeong~Joon Park, Peter Florence, Julian Straub, Richard Newcombe, and Steven
  Lovegrove.
\newblock {DeepSDF: Learning Continuous Signed Distance Functions for Shape
  Representation}.
\newblock In {\em The IEEE Conference on Computer Vision and Pattern
  Recognition}, 2019.

\bibitem{Paschalidou2019}
Despoina Paschalidou, Ali~Osman Ulusoy, and Andreas Geiger.
\newblock Superquadrics revisited: Learning 3d shape parsing beyond cuboids.
\newblock In {\em The IEEE Conference on Computer Vision and Pattern
  Recognition (CVPR)}, June 2019.

\bibitem{matryoshka}
Stephan~R. Richter and Stefan Roth.
\newblock {Matryoshka Networks: Predicting 3D Geometry via Nested Shape
  Layers}.
\newblock In {\em Proceedings IEEE Conf. on Computer Vision and Pattern
  Recognition (CVPR)}, 2018.

\bibitem{Soltani17}
Amir~Arsalan Soltani, Haibin Huang, Jiajun Wu, Tejas Kulkarni, and Joshua
  Tenenbaum.
\newblock Synthesizing 3d shapes via modeling multi-view depth maps and
  silhouettes with deep generative networks.
\newblock In {\em CVPR}, 2017.

\bibitem{pointnet}
Hao Su, Charles Qi, Kaichun Mo, and Leonidas Guibas.
\newblock {PointNet: Deep Learning on Point Sets for 3D Classification and
  Segmentation}.
\newblock In {\em CVPR}, 2017.

\bibitem{tatarchenko2017octree}
Maxim Tatarchenko, Alexey Dosovitskiy, and Thomas Brox.
\newblock Octree generating networks: Efficient convolutional architectures for
  high-resolution 3d outputs.
\newblock In {\em IEEE International Conference on Computer Vision (ICCV)},
  2017.

\bibitem{spheremesh}
Jean-Marc Thiery, Emilie Guy, and Tamy Boubekeur.
\newblock Sphere-meshes: Shape approximation using spherical quadric error
  metrics.
\newblock {\em ACM Transaction on Graphics (Proc. SIGGRAPH Asia 2013)},
  32(6):Art. No. 178, 2013.

\bibitem{animatedsm}
Jean-Marc Thiery, \'{E}milie Guy, Tamy Boubekeur, and Elmar Eisemann.
\newblock Animated mesh approximation with sphere-meshes.
\newblock {\em ACM Trans. Graph.}, pages 30:1--30:13, 2016.

\bibitem{smhandtrack}
Anastasia Tkach, Mark Pauly, and Andrea Tagliasacchi.
\newblock Sphere-meshes for real-time hand modeling and tracking.
\newblock {\em ACM Trans. Graph.}, 35(6), 2016.

\bibitem{Tulsiani2017}
Shubham Tulsiani, Hao Su, Leonidas~J. Guibas, Alexei~A. Efros, and Jitendra
  Malik.
\newblock Learning shape abstractions by assembling volumetric primitives.
\newblock In {\em The IEEE Conference on Computer Vision and Pattern
  Recognition (CVPR)}, July 2017.

\bibitem{pixel2mesh}
Nanyang Wang, Yinda Zhang, Zhuwen Li, Yanwei Fu, Wei Liu, and Yu-Gang Jiang.
\newblock Pixel2mesh: Generating 3d mesh models from single rgb images.
\newblock In {\em ECCV}, 2018.

\bibitem{Wang-2017-ocnn}
Peng-Shuai Wang, Yang Liu, Yu-Xiao Guo, Chun-Yu Sun, and Xin Tong.
\newblock {O-CNN: Octree-based Convolutional Neural Networks for 3D Shape
  Analysis}.
\newblock {\em ACM Transactions on Graphics (SIGGRAPH)}, 36(4), 2017.

\bibitem{obbcage}
Chuhua Xian, Hongwei Lin, and Shuming Gao.
\newblock Automatic cage generation by improved obbs for mesh deformation.
\newblock {\em The Visual Computer}, 28(1):21--33, 2012.

\bibitem{nnrotation}
Yi Zhou, Connelly Barnes, Lu Jingwan, Yang Jimei, and Li Hao.
\newblock On the continuity of rotation representations in neural networks.
\newblock In {\em The IEEE Conference on Computer Vision and Pattern
  Recognition (CVPR)}, June 2019.

\bibitem{gdc}
Yang Zhou, Kangxue Yin, Hui Huang, Hao Zhang, Minglun Gong, and Daniel
  Cohen-Or.
\newblock Generalized cylinder decomposition.
\newblock {\em ACM Trans. Graph.}, 34(6), 2015.

\bibitem{minimumncd}
{Zhou Ren}, {Junsong Yuan}, {Chunyuan Li}, and {Wenyu Liu}.
\newblock Minimum near-convex decomposition for robust shape representation.
\newblock In {\em 2011 International Conference on Computer Vision}, Nov 2011.

\end{thebibliography}
}

\clearpage
\includepdf[pages=1-last]{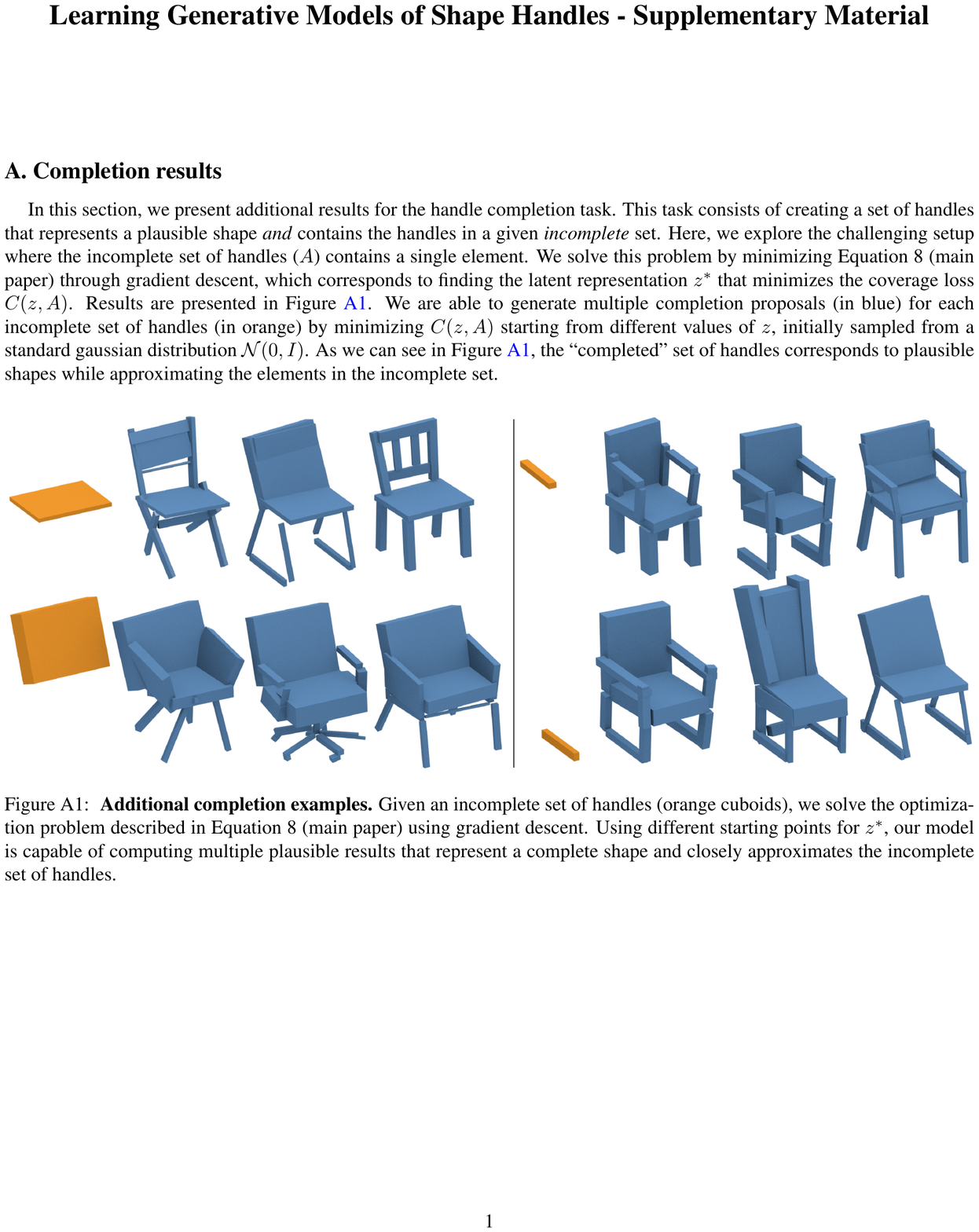}

\end{document}